\pdfoutput=1
\documentclass[11pt]{article}
\usepackage[margin=1in]{geometry}
\usepackage[T1]{fontenc}
\usepackage{lmodern}
\usepackage{microtype}
\usepackage{authblk}
\usepackage{caption}
\usepackage{booktabs}
\usepackage{cite}
\usepackage{amsmath,amssymb,amsfonts}
\usepackage{graphicx}
\usepackage{textcomp}
\usepackage[dvipsnames]{xcolor}
\usepackage[colorlinks=true,allcolors=blue!60!black,pdftitle={Auditing Path-Specific Fairness Across Causal Discovery Algorithms: A Heart Failure Case Study},pdfauthor={Nitish Nagesh, Elahe Khatibi, Thomas Dean Hughes, Mahdi Bagheri, Pratik Gajane, Amir M. Rahmani}]{hyperref}
\usepackage{subcaption}
\usepackage{tikz}
\usepackage{pgfplots}
\usepackage[utf8]{inputenc}
\usetikzlibrary{arrows.meta, positioning, calc}
\pgfplotsset{compat=1.18}

\begin{document}

\title{Auditing Path-Specific Fairness Across Causal Discovery Algorithms: A Heart Failure Case Study}

\author[1]{Nitish Nagesh}
\author[1]{Elahe Khatibi}
\author[1]{Thomas Dean Hughes}
\author[1]{Mahdi Bagheri}
\author[2]{Pratik Gajane}
\author[1]{Amir M. Rahmani}
\affil[1]{University of California, Irvine, Irvine, CA, USA\\
\texttt{\{nnagesh1, ekhatibi, thomasdh, mahdib1, amirr1\}@uci.edu}}
\affil[2]{University of Orl\'eans, Orl\'eans, France\\
\texttt{pratik.gajane@univ-orleans.fr}}
\date{}

\maketitle
\begin{abstract}
Causal discovery is increasingly applied to healthcare data, where the recovered relationships can encode biased pathways that propagate to downstream prediction. Path-specific fairness analysis, through direct, indirect, and spurious effect decompositions along with fairness--utility ratio assessment allows bias quantification and associated prediction cost. However, existing evaluations of causal discovery under fairness constraints have largely used non-clinical datasets and have not separated the contribution of confounders from that of mediators, leaving open how discovery-algorithm choice propagates through path-specific fairness to the utility trade-off in clinical settings. This work addresses that gap by establishing an expert-elicited causal graph benchmark for the heart failure clinical records dataset and tracing discovery algorithms through composite, per-variable fairness decompositions and fairness-utility ratio. 
ICA-LiNGAM achieves the lowest structural Hamming distance and highest recall, while GES achieves the highest precision. 
For path-specific effects, there was a 1.2 percentage-point difference in the mediated effect on ejection fraction on overall causal fairness between the original and discovered graphs. 
This ultimately led to variations in the fairness-utility ratio highlighting need for fine-grained evaluation.
These results indicate that discovery-aware, path-specific fairness evaluation is consequential in clinical settings and warrants inclusion in routine auditing of clinical risk models.
\end{abstract}


\section{Introduction}
Causal discovery is a tool for uncovering relationships between variables across multiple domains including healthcare\cite{brouillard_landscape_2025}. The discovered relationships have multiple facets with potential for biased pathways\cite{makhlouf2024causality}. Recent advances in fairness evaluation\cite{chiappa2019path,plecko2024causalfairnessanalysistoolkit} enables understanding effects of fairness pathways. Further, beyond performance evaluation, trade-offs between utility and fairness\cite{plecko2024reconcilingparity} help provide a nuanced understanding of biased pathways for potential intervention.
However, current work on evaluating causal discovery algorithm on fairness constraints\cite{binkyte2023causal,zanna2025fairness} focus on non-clinical datasets and fail to capture fairness effects of confounders on outcome. 
To this end, there is need to examine path-specific fairness across multiple causal discovery algorithms on clinical data along with examining trade-offs between utility and fairness.
To address these gaps, we benchmark a clinical dataset and apply a suite of causal discovery algorithms and analyze overall fairness, variable-level fairness across confounding and mediating paths and the causal fairness utility ratio. 

We make the following contributions:
\begin{itemize}
    \item We establish an expert-elicited causal graph benchmark for a heart failure clinical dataset.
    \item We map this benchmark to an existing fairness model and evaluate existing causal discovery algorithms against it.
    \item We carry out path-specific and per-variable analysis of the fairness-utility trade-off across discovered graph structures.
\end{itemize}

\section{Methodology}
\textbf{Overview.}
The evaluation framework, illustrated in Fig.~\ref{fig:proposed-architecture}, traces a single dataset through six stages.
The first stage involves performing causal structure discovery followed by evaluating composite and path-specific fairness for the benchmark and discovered graphs along with performance-fairness tradeoffs.

\begin{figure}[!t]
\centering
\resizebox{\textwidth}{!}{%
\begin{tikzpicture}[
    font=\scriptsize,
    every node/.style={align=center},
    >={Stealth[length=2mm,width=1.6mm]},
    procBox/.style={
        rectangle, draw, semithick, rounded corners=1.5pt,
        minimum width=26mm, minimum height=14mm,
        inner sep=2pt, fill=white
    },
    inBox/.style    ={procBox, fill=black!5,   draw=black!45},
    discBox/.style  ={procBox, fill=blue!7,    draw=blue!55!black},
    evalBox/.style  ={procBox, fill=orange!12, draw=orange!55!black},
    sfmBox/.style   ={procBox, fill=yellow!18, draw=olive!55!black,
                      minimum height=11mm, minimum width=32mm},
    decompBox/.style={procBox, fill=violet!10, draw=violet!55!black,
                      minimum height=11mm, minimum width=32mm},
    cfurBox/.style  ={procBox, fill=red!10,    draw=red!55!black,
                      minimum width=32mm},
    arr/.style={semithick, ->, draw=black!55},
    header/.style={font=\footnotesize\bfseries, anchor=south, text=black!75, text depth=0.5ex},
    currentOverlay/.style={
        draw=black!55, line width=0.9pt, dashed,
        rounded corners=4pt
    },
    novelOverlay/.style={
        draw=red!65!black, line width=1pt, dashed,
        rounded corners=4pt
    },
    currentTag/.style={
        font=\tiny\bfseries\sffamily, fill=black!55,
        text=white, rounded corners=1pt, inner sep=2.5pt
    },
    novelTag/.style={
        font=\tiny\bfseries\sffamily, fill=red!65!black,
        text=white, rounded corners=1pt, inner sep=2.5pt
    }
]
\node[header] at (0,    3.85) {1.~Inputs};
\node[header] at (3.4,  3.85) {2.~Causal\\Structure};
\node[header] at (6.8,  3.85) {3.~Structural\\Evaluation};
\node[header] at (10.6, 3.85) {4.~Fairness\\Mapping};
\node[header] at (14.4, 3.85) {5.~Fairness\\Decomposition};
\node[header] at (18.2, 3.85) {6.~Trade-off};

\node[inBox] (D) at (0, 2.2)  {Clinical dataset};
\node[inBox] (K) at (0, 0.4)  {Expert opinion +\\clinical literature};

\node[discBox] (Alg) at (3.4, 2.2)
    {Discovery algorithms\\\textbf{PC, FCI, GES,}\\\textbf{ICA-LiNGAM}};
\node[discBox] (Bench) at (3.4, 0.4)
    {Benchmark DAG};

\node[evalBox] (SE) at (6.8, 1.3)
    {Utility metrics\\SHD, TP/FP/FN/TN,\\Precision, Recall};

\node[sfmBox] (GS) at (10.6, 2.0)
    {\textbf{Benchmark graph}\\\textbf{fairness map}};
\node[sfmBox] (FS) at (10.6, 0.6)
    {\textbf{Discovered graph}\\\textbf{fairness map}};

\node[decompBox] (Comp) at (14.4, 2.0)
    {\textbf{Composite}\\\textbf{fairness}};
\node[decompBox] (PV)   at (14.4, 0.6)
    {\textbf{Per-variable}\\\textbf{fairness}};

\node[cfurBox] (CF) at (18.2, 1.3)
    {\textbf{Causal Fairness--}\\\textbf{Utility Ratio (CFUR)}};

\draw[arr] (D.east)     -- (Alg.west);
\draw[arr] (K.east)     -- (Bench.west);
\draw[arr] (Alg.east)   -- ($(SE.west)+(0,0.4)$);
\draw[arr] (Bench.east) -- ($(SE.west)+(0,-0.4)$);
\draw[arr] (SE.east)    -- ($(GS.west)+(0,-0.1)$);
\draw[arr] (SE.east)    -- ($(FS.west)+(0,0.1)$);
\draw[arr] (GS.east) to[out=0, in=180, looseness=0.6] (PV.west);
\draw[arr] (FS.east) to[out=0, in=180, looseness=0.6] (Comp.west);
\draw[arr] (Comp.east)  -- ($(CF.west)+(0,0.4)$);
\draw[arr] (PV.east)    -- ($(CF.west)+(0,-0.4)$);

\draw[currentOverlay]
    ($(D.north west)  +(-0.3, 0.55)$)
    rectangle
    ($(SE.south east) +(0.3, -0.7)$);
\node[currentTag, anchor=south west]
    at ($(D.north west)+(-0.25, 0.6)$)
    {Prior work};

\draw[novelOverlay]
    ($(GS.north west)+(-0.3, 0.85)$)
    rectangle
    ($(CF.south east)+(0.3, -1.4)$);
\node[novelTag, anchor=south west]
    at ($(GS.north west)+(-0.25, 0.9)$)
    {Beyond state-of-the-art};
\end{tikzpicture}%
}
\caption{Proposed methodology pipeline. The left overlay marks the prior-work pipeline: a clinical dataset and expert-elicited benchmark are passed through a suite of discovery algorithms (Stage~2), and the resulting structures are evaluated against the benchmark using utility metrics (Stage~3). The right overlay marks the contribution that extends beyond the state of the art. Both the benchmark and discovered graph is mapped to a Standard Fairness Model \cite{plecko2024causalfairnessanalysistoolkit} assignment of mediators and confounders (Stage~4); the total causal effect of the protected attribute on the outcome is then decomposed both at the composite level and per variable, exposing which channels and which covariates carry the bias under each graph (Stage~5); and the causal fairness--utility ratio (CFUR) is computed per channel under each map (Stage~6).}
\label{fig:proposed-architecture}
\end{figure}

\textbf{Dataset.}
\textcolor{black}{We use the heart failure clinical records dataset\cite{chicco2020machine} for evaluating causal discovery algorithms and path-specific effects. The dataset comprises 299 participants with a combination of demographic and clinical features and is open source making it well suited for analysis. The features include demographic variables such as age, gender, comorbidities such as anaemia, diabetes, hypertension status, smoking status, physiological measurements such as serum creatinine, serum sodium, ejection fraction, platelets and creatinine phosphokinase (CPK), time period patient was monitored starting from the time they were admitted to them passing away and the mortality outcome.}
 
\textbf{Ground Truth Causal Graph.}
\textcolor{black}{
We establish the causal graph benchmark shown in Fig.~\ref{fig:graph-comparison}a by working closely with a domain expert. 
}
\textcolor{black}{Higher blood pressure and elevated serum creatinine indicate presence of chronic kidney disease causing electrolyte imbalance leading to lower serum sodium\cite{tang2024evaluation}. Further, advanced heart failure increases likelihood of cardiac injury leading to elevated creatine phosphokinase (CPK) levels\cite{ali1999clinical}.
Lifestyle and metabolic factors act as critical precursors in this network. Smoking serves as a significant accelerator by causing elevated blood pressure and doubling the risk of developing heart failure \cite{ding2022cigarette}. Similarly, untreated hypertension eventually causes heart failure and a subsequent reduction in ejection fraction \cite{murphy2020heart}. Diabetes acts as a central node in this causal structure, directly leading to kidney disease and frequently co-existing with high blood pressure to exacerbate heart failure progression \cite{zhao2014blood}. Anemia often emerges as a consequence of CKD, forcing the heart to work harder to pump oxygen-deficient blood, thereby worsening heart failure outcomes \cite{silverberg2006anemia}. Serum creatinine levels may rise due to direct damage from diabetes or as a secondary effect of low ejection fraction, leading to fluid retention and decreased serum sodium \cite{silverberg2006anemia}. Finally, a lack of consistent follow-up, particularly in patients with chronic conditions like diabetes, leads to poor disease management and an increased probability of a death event \cite{ezekowitz2005impact}.
}

\textbf{Causal Discovery and evaluation.}
We perform causal discovery using four algorithms: PC (Peter Clark)\cite{spirtes2000causation,colombo2014order}, FCI (Fast Causal Inference)\cite{spirtes2013causal}, GES (Greedy Equivalence Search)\cite{chickering2002optimal}, and ICA-LiNGAM (Independent Component Analysis-Linear Non-Gaussian Acyclic Model)\cite{shimizu2006linear}. 
We run experiments for the aforementioned algorithms by leveraging Python implementation of \texttt{causal-learn}\cite{zheng2024causal}.

\textbf{Utility and fairness metrics.}
\textcolor{black}{We use different utility metrics including True Positive (TP), False Positive (FP), False Negative (FN), True Negative (TN) and Structural Hamming Distance (SHD).}
\textcolor{black}{Going beyond generic fairness, causal fairness provides a way to capture indirect effect to capture impact of mediators and spurious effect enabling understanding of effect of confounders.}
\textcolor{black}{We leverage the causal fairness utility ratio (CFUR)~\cite{plecko2025fairnessaccuracytradeoff} to evaluate the tension between fairness and accuracy.}

\textbf{Fairness Mapping, Decomposition, and Trade-off.}
The benchmark and discovered graph are mapped to a fairness model~\cite{plecko2024causalfairnessanalysistoolkit}, which maps covariates into mediators, confounders, sensitive attribute and outcome depending on the dataset and variables under consideration.
Each of these graphs are mapped to a composite fairness decomposition comprising direct, indirect, and spurious effects. 
Further, we decompose each covariate's individual composition.
Finally, the causal fairness--utility ratio (CFUR)~\cite{plecko2025fairnessaccuracytradeoff} is computed per path.
Analyzing the CFUR separately for the benchmark-aligned and discovered graphs allows examining granular examination of fairness-utility trade-offs.

\begin{figure}[t]
\centering
\begin{subfigure}[t]{0.48\linewidth}
\centering
\resizebox{\linewidth}{!}{%
\begin{tikzpicture}[
    >={Stealth[length=2.2mm,width=1.8mm]},
    var/.style={circle, draw, thick, minimum size=10mm,
        align=center, font=\tiny, inner sep=0.5pt, fill=white},
    mediator/.style  ={var, fill=blue!10},
    confounder/.style={var, fill=orange!25},
    protected/.style ={var, fill=red!15, draw=red!70!black, line width=0.8pt},
    outcome/.style   ={var, fill=gray!25, line width=0.8pt, double, double distance=0.8pt},
    direct/.style    ={ultra thick, -{Stealth[length=3mm,width=2.5mm]}, black},
    mediating/.style ={thick, dashed, -{Stealth[length=2.2mm,width=1.8mm]}, blue!65!black},
    confound/.style  ={thin, -{Stealth[length=2.2mm,width=1.8mm]}, orange!75!black},
    cyclic/.style    ={thick, {Stealth[length=2.2mm,width=1.8mm]}-{Stealth[length=2.2mm,width=1.8mm]}, teal!60!black}
]
\node[mediator]   (Smoke)  at (-4.0, 1.0)  {Smoking};
\node[protected]  (Gender) at (-4.0,-1.8)  {Gender};
\node[mediator]   (HBP)    at (-1.5, 1.0)  {High Blood\\Pressure};
\node[mediator]   (SS)     at (-0.5, 3.2)  {Serum\\Sodium};
\node[mediator]   (SC)     at ( 1.5, 0.0)  {Serum\\Creatinine};
\node[confounder] (Plt)    at ( 1.5, 3.2)  {Platelets};
\node[mediator]   (EF)     at ( 4.0, 1.5)  {Ejection\\Fraction};
\node[confounder] (Age)    at ( 4.0, 3.2)  {Age};
\node[confounder] (FT)     at ( 6.5, 3.2)  {Followup\\Time};
\node[outcome]    (Death)  at ( 6.5, 0.0)  {Death};
\node[mediator]   (Diab)   at (-1.5,-1.8)  {Diabetes};
\node[mediator]   (CPK)    at ( 1.5,-2.5)  {CPK};
\node[mediator]   (Ana)    at ( 4.0,-1.5)  {Anaemia};
\draw[direct] (Gender) to[bend left=8] (Death);
\draw[mediating] (Gender) -- (Smoke);
\draw[mediating] (Gender) -- (Diab);
\draw[mediating] (Smoke)  -- (HBP);
\draw[mediating] (Diab)   -- (HBP);
\draw[mediating] (Diab)   -- (SC);
\draw[mediating] (Diab)   -- (CPK);
\draw[mediating] (HBP)    -- (SC);
\draw[mediating] (HBP)    -- (EF);
\draw[mediating] (HBP)    to[bend right=18] (Death);
\draw[mediating] (SC)     -- (SS);
\draw[mediating] (SC)     -- (EF);
\draw[mediating] (SC)     -- (Death);
\draw[mediating] (CPK)    -- (Ana);
\draw[mediating] (Ana)    -- (EF);
\draw[mediating] (Ana)    -- (Death);
\draw[mediating] (EF)     -- (Death);
\draw[confound] (Age) -- (EF);
\draw[confound] (Age) -- (Death);
\draw[confound] (FT)  -- (Death);
\draw[confound] (EF)  -- (Plt);
\draw[cyclic] (SS) -- (EF);
\end{tikzpicture}%
}
\caption{Expert-derived benchmark.}
\label{fig:benchmark-graph}
\end{subfigure}\hfill
\begin{subfigure}[t]{0.48\linewidth}
\centering
\resizebox{\linewidth}{!}{%
\begin{tikzpicture}[
    >={Stealth[length=2.2mm,width=1.8mm]},
    var/.style={circle, draw, thick, minimum size=10mm,
        align=center, font=\tiny, inner sep=0.5pt, fill=white},
    mediator/.style    ={var, fill=blue!10},
    confounder/.style  ={var, fill=orange!25},
    protected/.style   ={var, fill=red!15, draw=red!70!black, line width=0.8pt},
    outcome/.style     ={var, fill=gray!25, line width=0.8pt, double, double distance=0.8pt},
    isoMed/.style      ={var, fill=blue!10, dash pattern=on 2pt off 1.5pt},
    isoConf/.style     ={var, fill=orange!25, dash pattern=on 2pt off 1.5pt},
    arr/.style         ={thick, -{Stealth[length=2.2mm,width=1.8mm]}, black},
    biarr/.style       ={thick, {Stealth[length=2.2mm,width=1.8mm]}-{Stealth[length=2.2mm,width=1.8mm]}, teal!60!black},
    circarr/.style     ={thick, {Circle[open,length=2mm,line width=0.5pt]}-{Stealth[length=2.2mm,width=1.8mm]}, black}
]
\node[mediator]   (Smoke)  at (-4.0, 1.0)  {Smoking};
\node[protected]  (Gender) at (-4.0,-1.8)  {Gender};
\node[mediator]   (HBP)    at (-1.5, 1.0)  {High Blood\\Pressure};
\node[mediator]   (SS)     at (-0.5, 3.2)  {Serum\\Sodium};
\node[mediator]   (SC)     at ( 1.5, 0.0)  {Serum\\Creatinine};
\node[isoConf]    (Plt)    at ( 1.5, 3.2)  {Platelets};
\node[mediator]   (EF)     at ( 4.0, 1.5)  {Ejection\\Fraction};
\node[confounder] (Age)    at ( 4.0, 3.2)  {Age};
\node[confounder] (FT)     at ( 6.5, 3.2)  {Followup\\Time};
\node[outcome]    (Death)  at ( 6.5, 0.0)  {Death};
\node[isoMed]     (Diab)   at (-1.5,-1.8)  {Diabetes};
\node[mediator]   (CPK)    at ( 1.5,-2.5)  {CPK};
\node[mediator]   (Ana)    at ( 4.0,-1.5)  {Anaemia};
\draw[circarr] (Smoke) -- (Gender);
\draw[biarr]   (SS)    -- (Age);
\draw[circarr] (SC)    -- (SS);
\draw[circarr] (SC)    to[bend right=18] (Death);
\draw[circarr] (SS)    to[bend left=28]  (Death);
\draw[circarr] (CPK)   -- (Ana);
\draw[arr]     (Ana)   -- (EF);
\draw[arr]     (FT)    -- (Death);
\draw[biarr]   (EF)    -- (Death);
\draw[circarr] (HBP)   to[bend left=25] (FT);
\draw[biarr]   (Ana)   to[bend right=35] (FT);
\end{tikzpicture}%
}
\caption{FCI-discovered PAG.}
\label{fig:fci-graph}
\end{subfigure}
\caption{Benchmark (a) vs.~FCI-discovered PAG (b). Node fills: mediator (blue), confounder (orange), protected attribute Gender (red), outcome Death (double-bordered gray). Dashed borders mark FCI-isolated nodes. Benchmark edges: bold (direct), dashed blue (mediating), thin orange (confounding), double-headed teal (cyclic). PAG edges: $A\!\to\!B$ (directed), $A\!\circ\!\!\to\!B$ ($B$ not an ancestor of $A$), $A\!\leftrightarrow\!B$ (latent common cause).}
\label{fig:graph-comparison}
\end{figure}

\section{Results}

\textbf{Structural recovery.}
As shown in Table~\ref{tab:adjacency-hfcr}, ICA-LiNGAM has lowest skeleton structural hamming distance and highest recall while GES has the highest precision and lowest number of false positives. PC and FCI produce the same adjacency matrices, neglecting the effect of unobserved confounders that FCI provides. 
Here skeleton structural hamming distance (SHD) counts only addition and deletion of edges without including the effect of edge reversals since we ignore directionality in our problem formulation.

\begin{table}[htbp]
\centering
\caption{Structural agreement of discovery algorithms with the benchmark.}
\label{tab:adjacency-hfcr}
\begin{tabular}{l|r|rrrr|rr}
\toprule
\textbf{Method} & \textbf{Skel.\ SHD} & \textbf{TP} & \textbf{FP} & \textbf{FN} & \textbf{TN} & \textbf{Prec} & \textbf{Rec} \\
\midrule
PC         & 24 &  7 &  4 & 20 &  47 & 0.636 & 0.259 \\
FCI        & 24 &  7 &  4 & 20 &  47 & 0.636 & 0.259 \\
GES        & 22 &  6 &  1 & 21 &  50 & 0.857 & 0.222 \\
ICA-LiNGAM & 20 & 10 &  3 & 17 &  48 & 0.769 & 0.370 \\
\bottomrule
\end{tabular}
\end{table}

\begin{figure}[htbp]
\centering
\begin{tikzpicture}
\begin{axis}[
    ybar, bar width=10pt, width=0.72\linewidth, height=0.36\linewidth,
    enlarge x limits=0.35, ymin=-50, ymax=40,
    ytick={-40,-20,0,20,40},
    ylabel={Effect (\%)}, xlabel={Causal Path},
    symbolic x coords={DE,IE,SE}, xtick=data,
    ymajorgrids=true, grid style={gray!25, dashed},
    axis y line*=left, axis x line*=bottom,
    legend style={at={(0.5,-0.25)}, anchor=north, legend columns=2,
        draw=none, column sep=0.4cm, font=\footnotesize},
    label style={font=\footnotesize}, tick label style={font=\footnotesize},
    error bars/error bar style={line width=0.7pt, black},
    error bars/error mark={|}]
\addplot+[fill=gray!25, draw=black, line width=0.5pt,
    error bars/.cd, y dir=both, y explicit]
    coordinates {(DE,-6.88) +- (0,39.50) (IE,-4.48) +- (0,40.31) (SE,-1.56) +- (0, 4.42)};
\addplot+[fill=black!65, draw=black, line width=0.5pt,
    error bars/.cd, y dir=both, y explicit]
    coordinates {(DE, 3.17) +- (0,6.02) (IE, 5.03) +- (0,5.66) (SE,-1.77) +- (0,3.45)};
\legend{Benchmark Graph, FCI-discovered Graph}
\end{axis}
\end{tikzpicture}
\caption{Composite direct (DE), indirect (IE), and spurious (SE) effects of the protected attribute on the outcome under the benchmark and FCI-discovered graphs. Bars show mean effect and the error bars span one standard deviation.}
\label{fig:composite-decomp}
\end{figure}

\textbf{Composite Fairness.}
The direct, indirect and spurious effects of the protected attribute on the outcome are shown in Fig.~\ref{fig:composite-decomp}.
The direct effect of the protected attribute on the outcome flipped sign between graphs, from $-6.88\%$ under the benchmark to $+3.17\%$ under the FCI-discovered graph, but both estimates are statistically indistinguishable from zero given the wide error bars under the benchmark.
The indirect effect likewise flipped sign, from $-4.48\%$ to $+5.03\%$, while the spurious effect remained small and negative in both.
For the benchmark graph, the standard deviations on direct and indirect effects exceeded $39\%$, rendering those composite estimates unreliable while they remained below $10\%$ for the FCI-discovered graph.

\textbf{Path-specific fairness decomposition.}
The per-variable contributions are shown in Fig.~\ref{fig:perv-decomp}. 
Ejection fraction is the dominant contributor to indirect effects.
Under the FCI-discovered graph, every covariate apart from ejection fraction was reclassified from an indirect to a spurious effect, and the per-variable direct effect flipped sign and shrunk an order of magnitude.
In the FCI-discovered graph, the indirect effect matches the composite (single mediator), while the spurious effect differs because the composite uses double machine learning whereas the individual decomposition uses parametric generalized linear models (GLMs) in a Monte Carlo simulation\cite{plecko2024reconcilingparity}. Variable order in the sequential naturalization affects how contributions are distributed across variables, and bootstrapping and sampling introduce additional variation in effect sizes.

\begin{figure}[htbp]
\centering
\begin{tikzpicture}[
    font=\footnotesize,
    bar/.style={draw=black!60, line width=0.25pt},
    se/.style ={bar, fill=orange!50},
    ie/.style ={bar, fill=blue!40},
    zeroLine/.style={black!70, line width=0.4pt},
    grid/.style={black!12, line width=0.2pt},
    rowLabel/.style={anchor=east, font=\footnotesize, text=black!85},
    sfmLabel/.style={font=\footnotesize\itshape, text=black!50}]
\def\sx{1.7}\def\barH{0.11}\def\inGap{0.14}\def\rowH{0.55}
\foreach \v in {-1,0,1,2,3,4}{
    \draw[grid] ({\v*\sx}, -0.05) -- ({\v*\sx}, {10*\rowH + 0.35});
    \node[font=\footnotesize, text=black!55, anchor=north] at ({\v*\sx}, -0.05) {\v};}
\draw[zeroLine] (0, -0.05) -- (0, {10*\rowH + 0.35});
\foreach \i/\name in {0/time, 1/smok., 2/SNa, 3/SCr, 4/plat.,
    5/HBP, 6/EF, 7/diab., 8/CPK, 9/anaem., 10/age}{
    \node[rowLabel] at ({-\sx-0.1}, {\i*\rowH}) {\name};}
\fill[se] (0,{10*\rowH + \inGap - \barH}) rectangle ({ 0.94*\sx},{10*\rowH + \inGap + \barH});
\fill[ie] (0,{ 9*\rowH + \inGap - \barH}) rectangle ({ 0.01*\sx},{ 9*\rowH + \inGap + \barH});
\fill[ie] (0,{ 8*\rowH + \inGap - \barH}) rectangle ({ 0.33*\sx},{ 8*\rowH + \inGap + \barH});
\fill[ie] (0,{ 7*\rowH + \inGap - \barH}) rectangle ({-0.28*\sx},{ 7*\rowH + \inGap + \barH});
\fill[ie] (0,{ 6*\rowH + \inGap - \barH}) rectangle ({ 3.59*\sx},{ 6*\rowH + \inGap + \barH});
\fill[ie] (0,{ 5*\rowH + \inGap - \barH}) rectangle ({ 0.15*\sx},{ 5*\rowH + \inGap + \barH});
\fill[ie] (0,{ 4*\rowH + \inGap - \barH}) rectangle ({ 0.38*\sx},{ 4*\rowH + \inGap + \barH});
\fill[ie] (0,{ 3*\rowH + \inGap - \barH}) rectangle ({-0.06*\sx},{ 3*\rowH + \inGap + \barH});
\fill[ie] (0,{ 2*\rowH + \inGap - \barH}) rectangle ({ 0.20*\sx},{ 2*\rowH + \inGap + \barH});
\fill[ie] (0,{ 1*\rowH + \inGap - \barH}) rectangle ({-0.08*\sx},{ 1*\rowH + \inGap + \barH});
\fill[ie] (0,{ 0*\rowH + \inGap - \barH}) rectangle ({ 0.04*\sx},{ 0*\rowH + \inGap + \barH});
\fill[se] (0,{10*\rowH - \inGap - \barH}) rectangle ({ 2.04*\sx},{10*\rowH - \inGap + \barH});
\fill[se] (0,{ 9*\rowH - \inGap - \barH}) rectangle ({-0.43*\sx},{ 9*\rowH - \inGap + \barH});
\fill[se] (0,{ 8*\rowH - \inGap - \barH}) rectangle ({ 0.37*\sx},{ 8*\rowH - \inGap + \barH});
\fill[se] (0,{ 7*\rowH - \inGap - \barH}) rectangle ({-0.31*\sx},{ 7*\rowH - \inGap + \barH});
\fill[ie] (0,{ 6*\rowH - \inGap - \barH}) rectangle ({ 2.41*\sx},{ 6*\rowH - \inGap + \barH});
\fill[se] (0,{ 5*\rowH - \inGap - \barH}) rectangle ({-0.58*\sx},{ 5*\rowH - \inGap + \barH});
\fill[se] (0,{ 4*\rowH - \inGap - \barH}) rectangle ({ 0.61*\sx},{ 4*\rowH - \inGap + \barH});
\fill[se] (0,{ 3*\rowH - \inGap - \barH}) rectangle ({-0.24*\sx},{ 3*\rowH - \inGap + \barH});
\fill[se] (0,{ 2*\rowH - \inGap - \barH}) rectangle ({ 0.45*\sx},{ 2*\rowH - \inGap + \barH});
\fill[se] (0,{ 1*\rowH - \inGap - \barH}) rectangle ({ 0.76*\sx},{ 1*\rowH - \inGap + \barH});
\fill[se] (0,{ 0*\rowH - \inGap - \barH}) rectangle ({ 0.33*\sx},{ 0*\rowH - \inGap + \barH});
\node[sfmLabel, anchor=west] at ({2.04*\sx + 0.08},{10*\rowH + \inGap}) {G};
\node[sfmLabel, anchor=west] at ({2.04*\sx + 0.08},{10*\rowH - \inGap}) {F};
\node[font=\footnotesize, text=black!70, anchor=north] at ({1.5*\sx}, -0.5) {Contribution (\%)};
\begin{scope}[shift={(0.6, -1.25)}]
\fill[ie] (0,0) rectangle (0.22, 0.16);
\node[anchor=west, font=\footnotesize] at (0.24, 0.08) {IE (mediator)};
\fill[se] (2.9,0) rectangle (3.12, 0.16);
\node[anchor=west, font=\footnotesize] at (3.14, 0.08) {SE (confounder)};
\end{scope}
\end{tikzpicture}
\caption{Per-variable contribution (\%) to the total effect of protected attribute on outcome. \emph{G}: Benchmark graph (upper bar); \emph{F}: FCI-discovered graph (lower bar). Ejection Fraction (EF) is the only variable preserved as a mediator across both graphs. Every other covariate flips from IE to SE. The direct effect is not represented here.}
\label{fig:perv-decomp}
\end{figure}

\begin{figure}[htbp]
\centering
\begin{tikzpicture}
\begin{axis}[
    ybar, bar width=10pt, width=0.72\linewidth, height=0.36\linewidth,
    enlarge x limits=0.35, ymin=-1.6, ymax=1.1,
    ytick={-1.5,-1,-0.5,0,0.5,1},
    ylabel={CFUR}, xlabel={Causal Path},
    symbolic x coords={DE,IE,SE}, xtick=data,
    ymajorgrids=true, grid style={gray!25, dashed},
    axis y line*=left, axis x line*=bottom,
    legend style={at={(0.5,-0.25)}, anchor=north, legend columns=2,
        draw=none, column sep=0.4cm, font=\footnotesize},
    label style={font=\footnotesize}, tick label style={font=\footnotesize},
    error bars/error bar style={line width=0.7pt, black},
    error bars/error mark={|}]
\addplot+[fill=gray!25, draw=black, line width=0.5pt,
    error bars/.cd, y dir=both, y explicit]
    coordinates {(DE,-0.30) +- (0,0.27) (IE, 0.40) +- (0,0.44) (SE, 0.35) +- (0,0.59)};
\addplot+[fill=black!65, draw=black, line width=0.5pt,
    error bars/.cd, y dir=both, y explicit]
    coordinates {(DE, 0.08) +- (0,0.02) (IE,-0.02) +- (0,0.08) (SE,-0.57) +- (0,0.86)};
\legend{Benchmark graph, FCI-discovered graph}
\end{axis}
\end{tikzpicture}
\caption{Causal Fairness Utility Ratio (CFUR) across direct (DE), indirect (IE), and spurious (SE) channels. Under the benchmark graph, IE and SE effects are positive and DE is negative while under the FCI-discovered graph, the SE channel flips strongly negative while DE and IE are near zero.}
\label{fig:cfur-comparison}
\end{figure}

\textbf{Causal Fairness--Utility Ratio.}
The causal fairness utility ratio (CFUR) ratio profiles differed both the ground truth graph that leverages the standard fairness map in compared to the FCI-discovered graph as shown in Fig.~\ref{fig:cfur-comparison}.
The former showcases positive ratio for indirect and spurious effect but negative ratio for direct effect. 
On the other hand, for the FCI-discovered graph, the spurious effect ratio switched to negative with direct and indirect effect ratio remaining close to zero. 
Large standard deviations on the indirect and spurious channels reflect high uncertainty. The same dataset thus yielded qualitatively different discovery pathways.

\section{Discussion}

We perform comprehensive evaluation of the different causal discovery methods on a clinical dataset. We analyze individual effects of different algorithms. We go one step further in disentangling path-specific effects. On the heart failure clinical records dataset, we observe the effect of ejection fraction, a significant contributor to patient death, reduced from $3.59\%$ in the benchmark graph to $2.41\%$ in the FCI-discovered graph. 
Further, there is an interchange between indirect and spurious path-specific effects coupled with inverted signs for causal-fairness utility ratio between the ground truth, underscoring the need for auditing causal discovery algorithms to better comprehend the intricate nuances of clinical datasets.

\textbf{Differentiation from state-of-the-art.} Our work extends prior fairness-aware causal discovery evaluations~\cite{binkyte2023causal,zanna2025fairness} to incorporate clinical data and distinguishes between contributions of mediators and confounders on individual and overall path-specific fairness, incorporating and extending literature~\cite{chiappa2019path,plecko2024causalfairnessanalysistoolkit}. A unique distinction is computing causal fairness utility ratio computed for mediating and confounding pathways beyond composite scores~\cite{plecko2025fairnessaccuracytradeoff}.

\textbf{Limitations.} 
The ground truth causal graph is based on insights from an individual domain expert coupled with information from literature.
Definitive causal directions are challenging to claim considering the graph is drawn after data is collected, thereby having a high propensity for correlation.
The benchmark also contains cyclic relationships violating the requirement of having directed acyclic graphs as is common place for traditional causal discovery algorithms.
Further, typical assumptions such as stationarity, faithfulness and identifiability may be violated.

\textbf{Future Work.} In the future, we intend to audit more causal discovery algorithms, examine multiple sensitive attributes, expand auditing fairness metrics across multiple sensitive attributes and increase the benchmark testing landscape. 
Taking into considerations time-series graphs, as is commonplace in real-world settings, helps expand the scope of discovery and fairness analysis.

\bibliographystyle{ieeetr}
\bibliography{ref}

\begin{thebibliography}{10}

\bibitem{brouillard_landscape_2025}
P.~Brouillard, C.~Squires, J.~Wahl, K.~P. Kording, K.~Sachs, A.~Drouin, and
  D.~Sridhar, ``The {Landscape} of {Causal} {Discovery} {Data}: {Grounding}
  {Causal} {Discovery} in {Real}-{World} {Applications},'' June 2025.
\newblock arXiv:2412.01953 [cs].

\bibitem{makhlouf2024causality}
K.~Makhlouf, S.~Zhioua, and C.~Palamidessi, ``When causality meets fairness: A
  survey,'' {\em Journal of Logical and Algebraic Methods in Programming},
  vol.~141, p.~101000, 2024.

\bibitem{chiappa2019path}
S.~Chiappa, ``Path-specific counterfactual fairness,'' in {\em Proceedings of
  the AAAI conference on artificial intelligence}, vol.~33, pp.~7801--7808,
  2019.

\bibitem{plecko2024causalfairnessanalysistoolkit}
D.~Plecko and E.~Bareinboim, ``Causal fairness analysis: A causal toolkit for
  fair machine learning,'' {\em Foundations and Trends in Machine Learning},
  vol.~17, no.~3, pp.~304--589, 2024.

\bibitem{plecko2024reconcilingparity}
D.~Plecko and E.~Bareinboim, ``Reconciling predictive and statistical parity: A
  causal approach,'' in {\em Proceedings of the AAAI Conference on Artificial
  Intelligence}, vol.~38, pp.~14625--14632, 2024.

\bibitem{binkyte2023causal}
R.~Binkyt{\.e}, K.~Makhlouf, C.~Pinz{\'o}n, S.~Zhioua, and C.~Palamidessi,
  ``Causal discovery for fairness,'' in {\em Workshop on Algorithmic Fairness
  through the Lens of Causality and Privacy}, pp.~7--22, PMLR, 2023.

\bibitem{zanna2025fairness}
K.~Zanna and A.~Sano, ``Fairness-driven llm-based causal discovery with active
  learning and dynamic scoring,'' {\em arXiv preprint arXiv:2503.17569}, 2025.

\bibitem{chicco2020machine}
D.~Chicco and G.~Jurman, ``Machine learning can predict survival of patients
  with heart failure from serum creatinine and ejection fraction alone,'' {\em
  BMC medical informatics and decision making}, vol.~20, no.~1, p.~16, 2020.

\bibitem{tang2024evaluation}
W.~W. Tang, M.~A. Bakitas, X.~S. Cheng, J.~C. Fang, S.~E. Fedson, A.~G.
  Fiedler, P.~Martens, W.~I. Mccallum, M.~O. Ogunniyi, J.~Rangaswami, {\em
  et~al.}, ``Evaluation and management of kidney dysfunction in advanced heart
  failure: a scientific statement from the american heart association,'' {\em
  Circulation}, vol.~150, no.~16, pp.~e280--e295, 2024.

\bibitem{ali1999clinical}
A.~S. Ali, B.~A. Rybicki, M.~Alam, N.~Wulbrecht, K.~Richer-Cornish, F.~Khaja,
  H.~N. Sabbah, and S.~Goldstein, ``Clinical predictors of heart failure in
  patients with first acute myocardial infarction,'' {\em American heart
  journal}, vol.~138, no.~6, pp.~1133--1139, 1999.

\bibitem{ding2022cigarette}
N.~Ding, A.~M. Shah, M.~J. Blaha, P.~P. Chang, W.~D. Rosamond, and
  K.~Matsushita, ``Cigarette smoking, cessation, and risk of heart failure with
  preserved and reduced ejection fraction,'' {\em Journal of the American
  College of Cardiology}, vol.~79, no.~23, pp.~2298--2305, 2022.

\bibitem{murphy2020heart}
S.~P. Murphy, N.~E. Ibrahim, and J.~L. Januzzi~Jr, ``Heart failure with reduced
  ejection fraction: a review,'' {\em Jama}, vol.~324, no.~5, pp.~488--504,
  2020.

\bibitem{zhao2014blood}
W.~Zhao, P.~T. Katzmarzyk, R.~Horswell, W.~Li, Y.~Wang, J.~Johnson, S.~B.
  Heymsfield, W.~T. Cefalu, D.~H. Ryan, and G.~Hu, ``Blood pressure and heart
  failure risk among diabetic patients,'' {\em International journal of
  cardiology}, vol.~176, no.~1, pp.~125--132, 2014.

\bibitem{silverberg2006anemia}
D.~S. Silverberg, D.~Wexler, A.~Iaina, S.~Steinbruch, Y.~Wollman, and
  D.~Schwartz, ``Anemia, chronic renal disease and congestive heart
  failure—the cardio renal anemia syndrome: the need for cooperation between
  cardiologists and nephrologists,'' {\em International urology and
  nephrology}, vol.~38, no.~2, pp.~295--310, 2006.

\bibitem{ezekowitz2005impact}
J.~A. Ezekowitz, C.~Van~Walraven, F.~A. McAlister, P.~W. Armstrong, and
  P.~Kaul, ``Impact of specialist follow-up in outpatients with congestive
  heart failure,'' {\em Cmaj}, vol.~172, no.~2, pp.~189--194, 2005.

\bibitem{spirtes2000causation}
P.~Spirtes, C.~N. Glymour, and R.~Scheines, {\em Causation, prediction, and
  search}.
\newblock MIT press, 2000.

\bibitem{colombo2014order}
D.~Colombo, M.~H. Maathuis, {\em et~al.}, ``Order-independent constraint-based
  causal structure learning.,'' {\em J. Mach. Learn. Res.}, vol.~15, no.~1,
  pp.~3741--3782, 2014.

\bibitem{spirtes2013causal}
P.~L. Spirtes, C.~Meek, and T.~S. Richardson, ``Causal inference in the
  presence of latent variables and selection bias,'' {\em arXiv preprint
  arXiv:1302.4983}, 2013.

\bibitem{chickering2002optimal}
D.~M. Chickering, ``Optimal structure identification with greedy search,'' {\em
  Journal of machine learning research}, vol.~3, no.~Nov, pp.~507--554, 2002.

\bibitem{shimizu2006linear}
S.~Shimizu, P.~O. Hoyer, A.~Hyv{\"a}rinen, A.~Kerminen, and M.~Jordan, ``A
  linear non-gaussian acyclic model for causal discovery.,'' {\em Journal of
  Machine Learning Research}, vol.~7, no.~10, 2006.

\bibitem{zheng2024causal}
Y.~Zheng, B.~Huang, W.~Chen, J.~Ramsey, M.~Gong, R.~Cai, S.~Shimizu,
  P.~Spirtes, and K.~Zhang, ``Causal-learn: Causal discovery in python,'' {\em
  Journal of Machine Learning Research}, vol.~25, no.~60, pp.~1--8, 2024.

\bibitem{plecko2025fairnessaccuracytradeoff}
D.~Plecko and E.~Bareinboim, ``Fairness-accuracy trade-offs: A causal
  perspective,'' in {\em Proceedings of the AAAI Conference on Artificial
  Intelligence}, vol.~39, pp.~26344--26353, 2025.

\end{thebibliography}

\end{document}